\documentclass[10pt,twocolumn,letterpaper]{article}

\usepackage{iccv}              

\usepackage{graphicx}
\usepackage{booktabs}
\usepackage{multirow}
\usepackage[table,xcdraw]{xcolor}
\usepackage{xcolor}
\usepackage{makecell}
\usepackage{pifont}
\usepackage{float}

\definecolor{iccvblue}{rgb}{0.21,0.49,0.74}
\usepackage[pagebackref,breaklinks,colorlinks,allcolors=iccvblue]{hyperref}

\def\paperID{10899} 
\def\confName{ICCV}
\def\confYear{2025}

\title{Dense2MoE: Restructuring Diffusion Transformer to MoE for Efficient Text-to-Image Generation}

\author{
Youwei Zheng\textsuperscript{1,2}\footnotemark[1] \quad
Yuxi Ren\textsuperscript{3} \quad
Xin Xia\textsuperscript{3} \quad
Xuefeng Xiao\textsuperscript{3} \quad
Xiaohua Xie\textsuperscript{1,4,5} \\
\textsuperscript{1}Sun Yat-sen University \quad
\textsuperscript{2}ByteDance Intelligent Creation \quad
\textsuperscript{3}ByteDance Seed Vision \\
\textsuperscript{4}Guangdong Province Key Laboratory of Information Security Technology \quad
\textsuperscript{5}Pazhou Lab (Huangpu) \\
\small \texttt{zhengyw33@mail2.sysu.edu.cn, \{renyuxi.20190622, xiaxin.97, xiaoxuefeng.ailab\}@bytedance.com} \\ 
\small \texttt{xiexiaoh6@mail.sysu.edu.cn}
}

\begin{document}

\twocolumn[{%
\maketitle
\vspace{-2.0em} 
\begin{figure}[H]
\hsize=\textwidth
\centering
\includegraphics[width=2.1\linewidth]{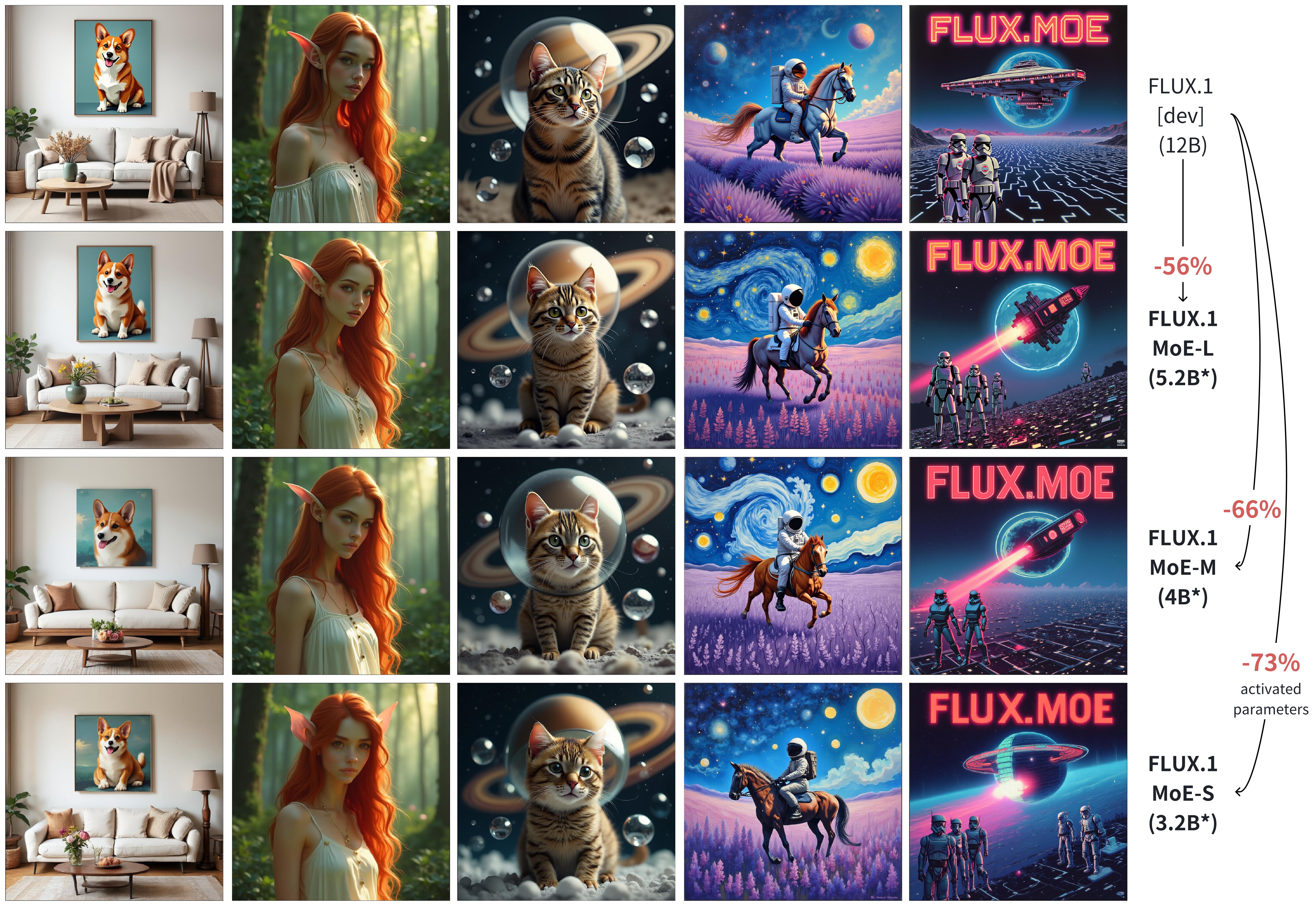}
\caption{The visual comparison between the 12B FLUX.1 [dev] and our FLUX.1-MoE models. The second, third, and fourth rows correspond to FLUX.1-MoE-L, FLUX.1-MoE-M, and FLUX.1-MoE-S, with 5.2B, 4B, and 3.2B activated parameters, respectively. These are sparse MoE models distilled from FLUX.1 [dev]. All images in each column are generated from the same random noise.}
\end{figure}
}]


\footnotetext[1]{Work done during internship at ByteDance Intelligent Creation.}

\begin{abstract}
Diffusion Transformer (DiT) has demonstrated remarkable performance in text-to-image generation; however, its large parameter size results in substantial inference overhead. Existing parameter compression methods primarily focus on pruning, but aggressive pruning often leads to severe performance degradation due to reduced model capacity. 
To address this limitation, we pioneer the transformation of a dense DiT into a Mixture of Experts (MoE) for structured sparsification, reducing the number of activated parameters while preserving model capacity.  
Specifically, we replace the Feed-Forward Networks (FFNs) in DiT Blocks with MoE layers, reducing the number of activated parameters in the FFNs by 62.5\%.
Furthermore, we propose the Mixture of Blocks (MoB) to selectively activate DiT blocks, thereby further enhancing sparsity.
To ensure an effective dense-to-MoE conversion, we design a multi-step distillation pipeline, incorporating Taylor metric-based expert initialization, knowledge distillation with load balancing, and group feature loss for MoB optimization.  
We transform large diffusion transformers (e.g., FLUX.1 [dev]) into an MoE structure, reducing activated parameters by 60\% while maintaining original performance and surpassing pruning-based approaches in extensive experiments. 
Overall, Dense2MoE establishes a new paradigm for efficient text-to-image generation.
\end{abstract}

\section{Introduction}
\label{sec:intro}

Diffusion models (DMs)~\cite{ho2020denoising, rombach2022high} have shown remarkable performance in generative tasks. 
Recently, there has been a trend that adopts the diffusion transformer (DiT) architecture~\cite{peebles2023scalable} for DMs, with continuously increasing model size. 
For example, FLUX.1~\cite{flux2024}, an advanced text-to-image model based on DiT, has 12 billion parameters, making it 13.8 times larger than SD1.5~\cite{rombach2022high}. 
However, as the models scale up, memory consumption and inference time increase, creating challenges for users due to the high cost.

Extensive research is conducted into efficient diffusion models to improve computational efficiency. Key topics of focus include advancements in faster sampler~\cite{song2020denoising, lu2022dpm}, step distillation~\cite{luo2023latent, ren2024hypersd, yin2024improved}, and model compression~\cite{fang2023structural, kim2023bksdm}, among others.
In this work, we focus on model compression, which we define as \textit{selecting a subset of parameters from the original model}.
Pruning techniques aim to identify an optimal subset of parameters by assessing the importance of weights. 
However, these techniques often suffer from severe performance degradation under high compression rates, as reducing the total number of parameters inherently limits the \textit{model's capacity}, making it difficult to fully recover performance even with retraining.

\begin{figure}[h]
\includegraphics[width=1.0\linewidth]{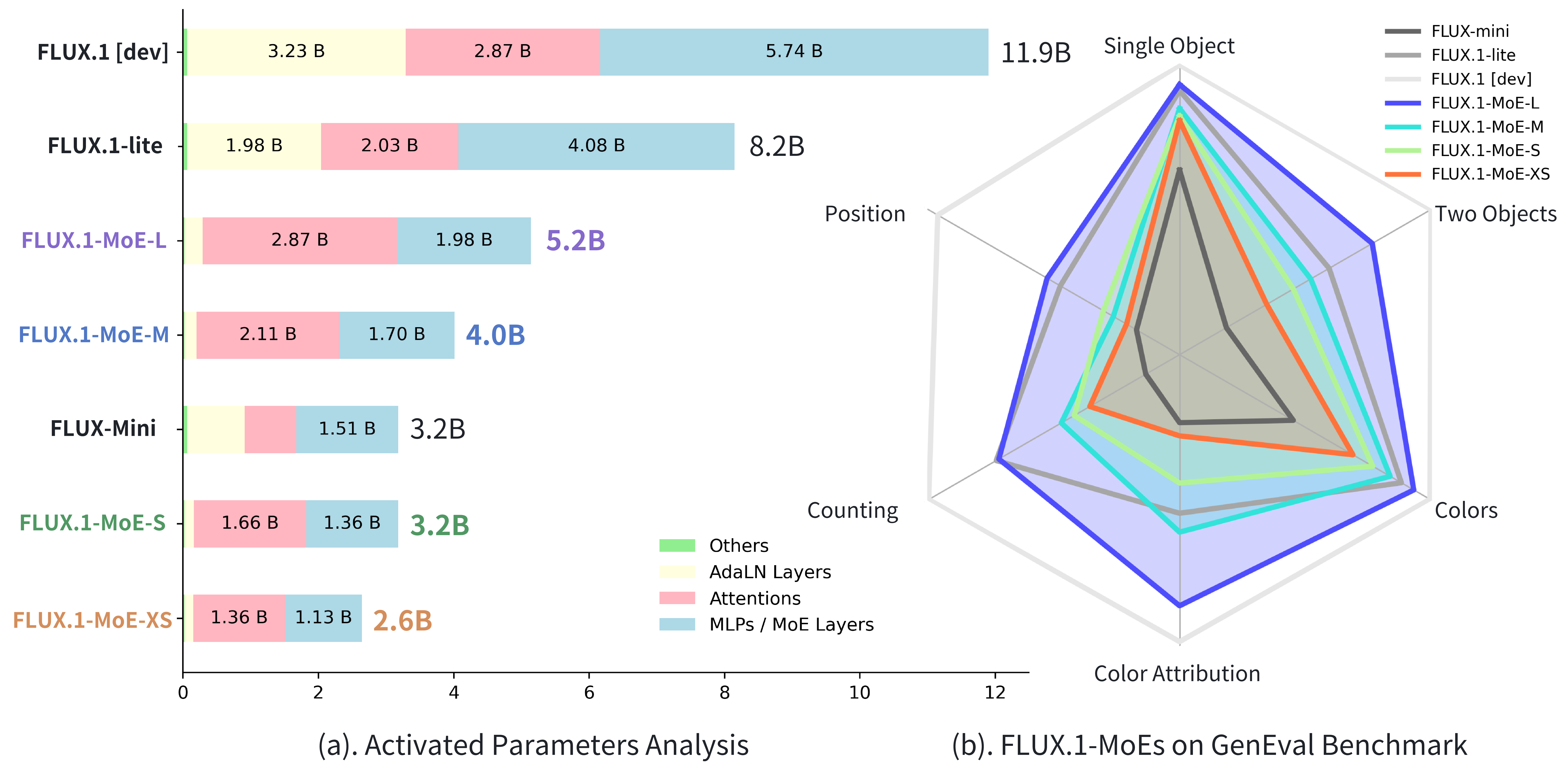}
\caption{Comparison of activated parameters and performance between FLUX.1-MoEs and the baseline models: The left figure shows the activated parameters, as well as the parameter count of the major modules in these models. The right figure compares our MoEs with the baseline on the GenEval~\cite{ghosh2023geneval} benchmark.}
\label{tab:head}
\end{figure}
\begin{figure}[h]
\includegraphics[width=1.0\linewidth]{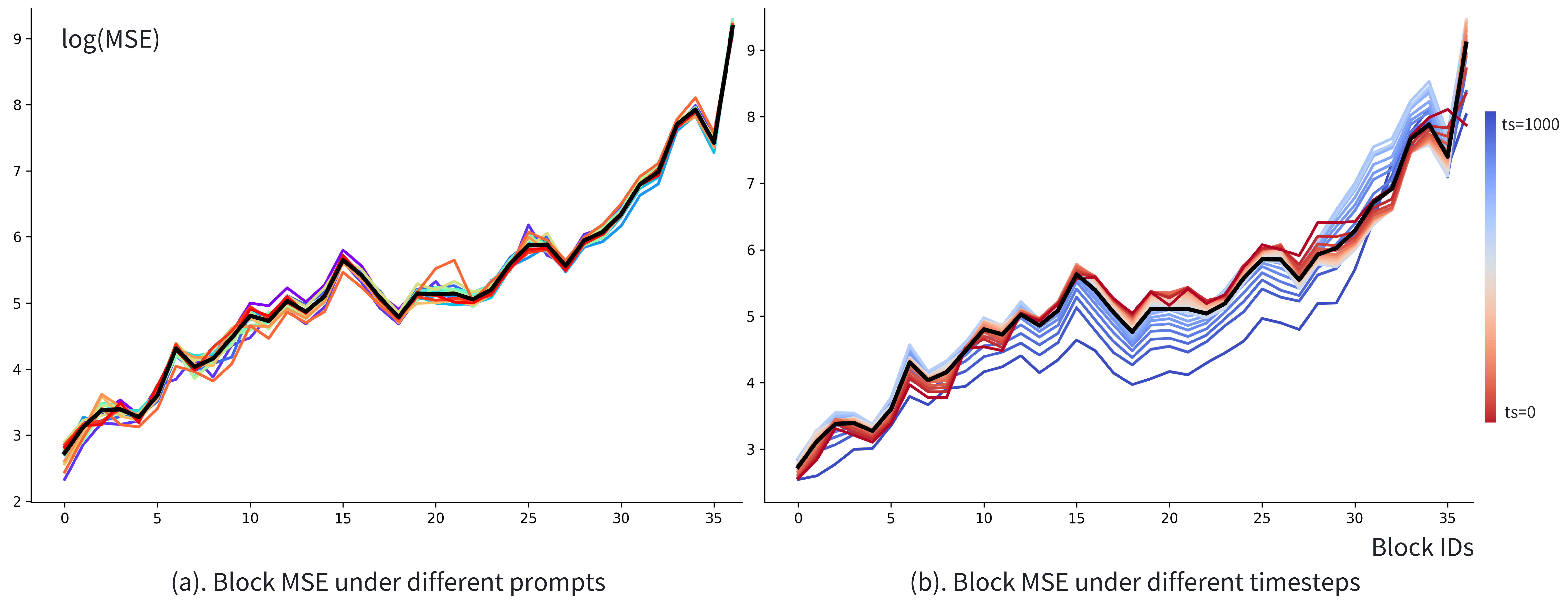}
\caption{Visualization of the MSE between the input and output of each single stream block in FLUX.1 [dev] under different prompts (left) and timesteps (right). The black line represents the average block MSE. The logarithm of the MSE is taken for better visualization.}
\label{tab:mse_vis}
\end{figure}

Rather than confining model compression to a fixed subset of parameters, we explore dynamically selecting different parameter subsets based on the input.
Ideally, this approach enables reducing the number of activated parameters without compromising model capacity.
Drawing inspiration from the Mixture of Experts (MoE)~\cite{shazeer2017outrageously} architecture, which has garnered significant attention in the LLM domain,
we propose transforming the dense DiT architecture into an MoE of the same size.
By allowing different inputs to activate distinct parameter subsets, our method effectively reduces computational cost.
More importantly, as illustrated in Fig.~\ref{tab:head}, FFNs in DiT account for nearly 50\% of the total parameters,
making MoE conversion a promising strategy for reducing activated parameters.

Besides the FFNs, we turn our attention to the blocks in DiTs.
We first compute the mean squared error (MSE) between transformer block inputs and outputs to assess block importance, particularly across varying timesteps and prompts at Fig.~\ref{tab:mse_vis}.
Notably, a block’s contribution varies significantly across timesteps and prompts. However, existing depth pruning methods only estimate block importance on average, neglecting sample-specific variations, potentially causing unrecoverable degradation.
Since text-to-image generation in DMs is a multi-step conditional denoising process, selectively activating the most critical blocks based on input could enhance efficiency without diminishing the model's expressiveness.

Based on these observations, we propose a structured sparsification framework for diffusion transformers, aiming to distill them into concise and efficient sparse architectures.
Firstly, we replace FFNs with MoE layers to reduce the activated parameter within each block. 
Secondly, we propose the Mixture of Blocks (MoB), a MoE-inspired structure that selectively activates blocks based on input features.
Additionally, we develop a dense-to-MoE pipeline, by first leveraging Taylor-metric and knowledge distillation for enhanced MoE initialization, applying a load-balancing loss for MoE distillation, and finally incorporating specialized group feature loss for MoB distillation. 
Using FLUX.1 [dev] as the base model, we achieve hybrid compression of activated parameters purely through knowledge distillation, reducing them from 12B to between 5.2B and 2.6B while maintaining performance and surpassing pruning methods of the same scale across multiple benchmarks.

Our contributions are as follows:
\begin{itemize}
    \item To the best of our knowledge, we present the first attempt at applying a dense-to-MoE paradigm to diffusion models, significantly reducing activated parameters while preserving model capacity.
    \item We propose a unified framework for structured sparsification, integrating the Mixture of Experts (MoE) and the Mixture of Blocks (MoB) in model design, along with a specialized knowledge distillation pipeline.
    \item We present FLUX.1-MoE, the first text-to-image model that transitions from a dense architecture to MoE. With a reduction of over 56\% in activation parameters, this model outperforms pruning-based methods while maintaining performance on par with FLUX.1 [dev].
\end{itemize}

\section{Related Works}
\textbf{Efficient Diffusion Models.}
Current research on efficient diffusion models encompasses acceleration techniques for sampling steps, such as sampler optimization~\cite{song2020denoising, lu2022dpm, luo2023latent}, step  distillation~\cite{ren2024hypersd,yin2024improved}, as well as model compression based on pruning~\cite{fang2023structural, kim2023bksdm, zhang2024laptopdiff, Lee2024koala, gupta2024progressive, flux1-lite, fang2024tinyfusion} or quantization~\cite{yang2024158bitflux}.
In particular, pruning, as the key approach to model parameter compression, typically involves analyzing the importance of weights to reduce parameters, followed by retraining or distillation to recover the model's performance. However, while pruning effectively reduces the parameter count, it inherently constrains the model’s overall capacity.
Moreover, there are impressive approaches~\cite{zhao2024dynamicdiffusiontransformer, ganjdanesh2025aptp} that dynamically reduce activated parameters during inference through carefully designed timestep-wise and token-wise masking mechanisms or prompt routers.
However, these methods come with complex and customized designs.
To achieve greater generality, we consider the mixture-of-experts (MoE) paradigm. MoE offers a simple yet powerful solution, integrating temporal, spatial, and text adaptability into a unified, efficient design. 
\\[0.5em]
\textbf{Dense to Mixture of Experts.}
In large language models~\cite{jiang2024mixtralexperts, deepseekai2024deepseekv3technicalreport, qwen25}, MoE architectures effectively scale model parameters while maintaining inference efficiency through sparse activation. 
Other sparse architectures, such as MoD~\cite{raposo2024mixtureofdepths}, introduce dynamic depth by allowing tokens to adaptively skip blocks.
Meanwhile, several methods~\cite{chen2023sparse, wu2024parameter, zhu2024moe} explore the Dense-to-MoE transition and repurpose the dense checkpoint as initial weights for MoE models.
In diffusion models, MoE has also been explored, with RAPHAEL~\cite{xue2023raphael} unifying temporal and spatial information in a mixture-of-experts architecture for text-to-image generation, and DiT-MoE~\cite{FeiDiTMoE2024} successfully scaling the DiT to 16B parameters. 
Inspired by these methods, we develop our dense-to-MoE pipeline for efficient diffusion models.
\section{Methods}

\begin{figure*}[h]
\hsize=\textwidth
\centering
\includegraphics[width=1.0\linewidth]{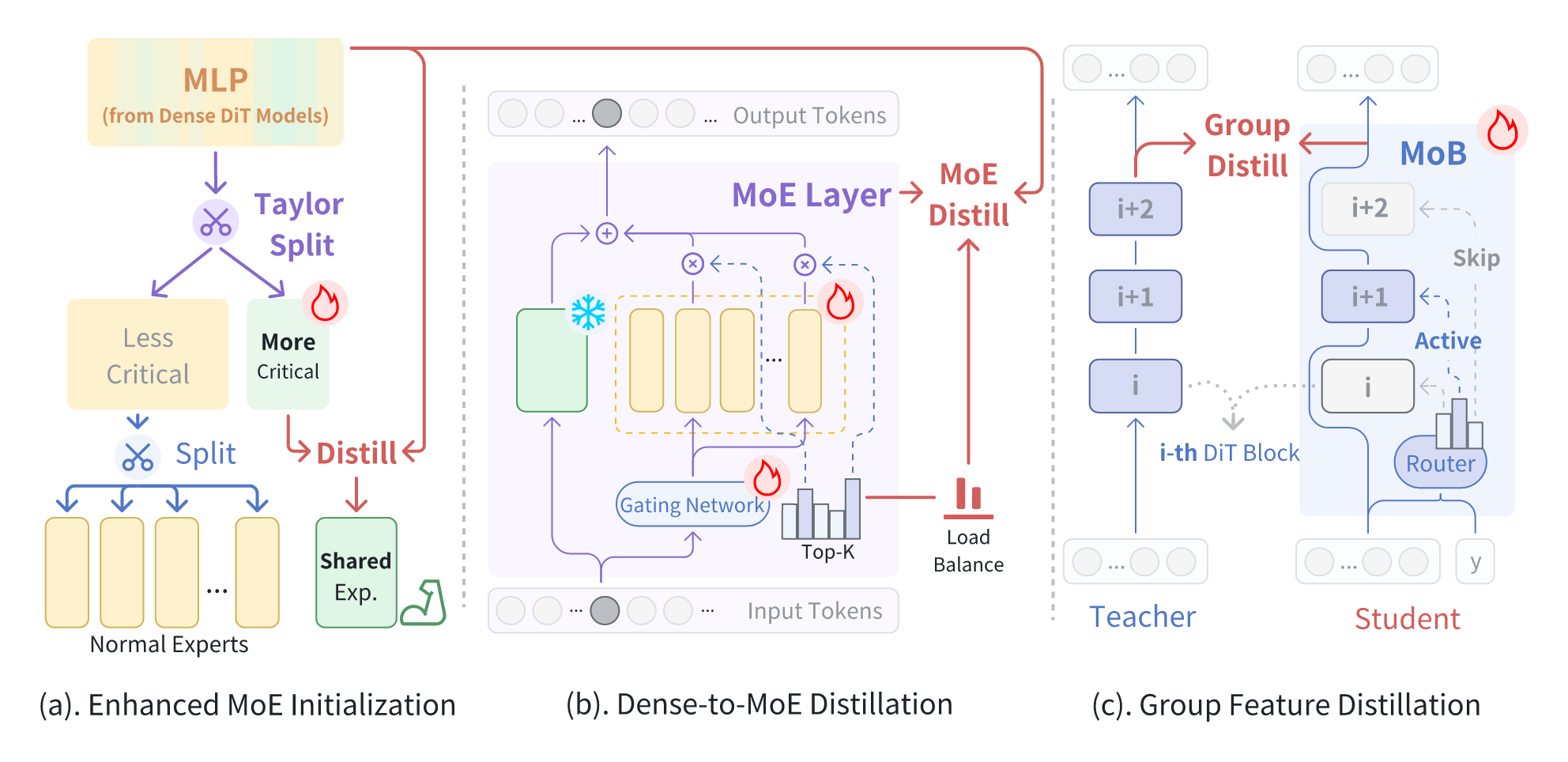}
\caption{The framework of Dense2MoE. The purple region represents the MoE layer, while the blue region denotes a MoB group. The pipeline comprises three stages: (a) Enhanced MoE Initialization, where MLP layers in DiT are restructured using a Taylor-based metric and knowledge distillation; (b) Dense-to-MoE Distillation, where assembles the enhanced weights into MoE and applies knowledge distillation with load balancing; (c) Group Feature Distillation for MoB, where blocks are grouped into MoB to further compress activated parameters by depth, with group features guiding the distillation.}
\label{fig:framework}
\end{figure*}

Our dense-to-MoE paradigm is integrated into both the model design and the distillation pipeline. In Sec.~\ref{subsec:sparse_model_design}, we introduce a hybrid sparse structure transformation that combines MoEs and MoBs. Subsequently, in Sec.~\ref{subsec:distillation_pipeline}, we present a distillation pipeline to further enhance the performance. 
For completeness, we provide the preliminaries of DMs and DiT architecture in the Appendix.

\subsection{Model Designs of Dense2MoE}
\label{subsec:sparse_model_design}

\textbf{Replacing FFNs with MoE Layers.}  
In the DiT, the feed-forward network (FFN) consists of a two-layer MLP with an activation function (e.g., GeLU) in between:  
\begin{equation}
\text{MLP}(x) = W_2( \text{GeLU}(W_1(x) + b_1)) + b_2.
\end{equation}  
The expansion ratio \( r \) of MLP determines the ratio between the input and hidden layer dimensions. Typically set to 4 in FFNs, means the hidden dimension is expanded to four times that of the input. Accordingly, the parameter count of an MLP layer is \(2 \times r \times h^2\), where \(h\) represents the input token's dimension.
To reduce the number of activated parameters, we replace the FFN with the MoE layer consisting of a gating network and two types of experts: (i) a shared expert and (ii) multiple standard experts. 
All the experts are also MLPs but are much smaller.
Let \( r_{\text{s}} \) denote the expansion ratio of the shared expert, \( r_{\text{n}} \) that of each normal expert, and \( n \) the total number of normal experts in the MoE layer. To maintain the same total expansion ratio \( r \) as the dense model, they are constrained by \(r = r_{\text{s}} + n \cdot r_{\text{n}}.\)
During inference, only the top \( k \) normal experts are activated for each token, resulting in an activated expansion ratio of \( r_a = r_{\text{s}} + k \cdot r_{\text{n}} \). Since the gating network has negligible parameter overhead, we define the activated parameter compression rate in FFN as \(\frac{r_a}{r}\).

In the MoE layer, the forward procedure for the \(t\)-th token is computed as follows:
\begin{align}
y^{(t)} = \text{MLP}_{\text{s}}(x^{(t)}) + \sum_{i=1}^{k} g(x^{(t)}, i) \cdot \text{MLP}_{\text{n}}^{(i)}(x^{(t)}), \\
g(x^{(t)}, i) = \begin{cases} 
{\alpha_t[i]} / {\sum_{j \in I^{(t)}} \alpha_t[j]}, & \text{if } i \in I^{(t)}, \\
0, & \text{otherwise},
\end{cases}
\end{align}
where \(\text{MLP}_{\text{s}}\) is the shared expert, \(\text{MLP}_{\text{n}}^{(i)}\) is the \(i\)-th selected expert, and \(g(x^{(t)}, i)\) is the gating output for the \(t\)-th token at the \(i\)-th selected expert. 
The gating network computes the gating output by first calculating the product of the gating weights and the input, then applying the softmax to it as \( \alpha_t = \text{softmax}(W_g x^{(t)}) \), selecting the top \(k\) experts \(I_t = \text{TopK}(\alpha_t, k)\), and finally normalizing it among the activated experts, setting all others to 0.
\\[0.5em]
\textbf{Grouping Blocks into Mixture of Blocks.}
Instead of typical MoE performed at the token level, we propose the Mixture of Blocks (MoB) to restructure transformer blocks and route at the feature level, dynamically selecting fewer blocks during inference.
MoB is well-suited for the diffusion sampling process, as it allows each sample to dynamically select which blocks to traverse at different timesteps and prompts.
In our implementation, consecutive transformer blocks are grouped into the MoB group.
The MoB group with \( m \) blocks consists of blocks from \( \textbf{B}^{(p)} \) to \( \textbf{B}^{(p+m)} \), where \( \kappa \) represents the number of activated blocks in the group. The block router selects blocks from \( \textbf{B}^{(q)} \) to \( \textbf{B}^{(q+\kappa)} \), subject to the constraints \( p \le q \) and \( q + \kappa \le p + m \). 
The forward pass of the MoB group is formulated as:
\begin{align}
&(x_t^{(p+m+1)}, c^{(p+m+1)}) = \notag \\ 
& \quad \textbf{B}^{(q+\kappa)} \left( \textbf{B}^{(q+\kappa-1)} \left( \cdots \left( \textbf{B}^{(q)}(x_t^{(p)},c^{(p)}) \right) \cdots \right) \right).
\end{align}
Here, \( x_t^{(p+m)} \) and \( c^{(p+m)} \) represent the image and text input features at the \( (p+m) \)-th block layer, and global condition embedding \( y \) for AdaLN is omitted for simplicity.
The MoB achieves a significant compression of activated parameters, as it can directly reduce \( m - \kappa \) blocks during inference, dynamically reducing the model's depth.

Moreover, conditions such as text and timestep are embedded into the global embedding \( y \) for AdaLN modulation in DiT blocks. Based on this, we develop a joint routing mechanism, reusing \( y \) to enable the router to explicitly capture these conditions. The routing in MoB is formulated as
\begin{equation}
\text{TopK}(\alpha W_x([x^{(p)}, c^{(p)}]) + (1-\alpha)W_yy, \kappa),
\end{equation}
where \( W_x \) and \( W_y \) are the gating weights for the concatenated input features \(x^{(p)}, c^{(p)}\), and \( \alpha \) is the weight.

\subsection{Distillation Pipeline of Dense2MoE }
\label{subsec:distillation_pipeline}
As shown in Fig.~\ref{fig:framework}, we design a distillation pipeline for MoE and MoB transformation. Since MoE and MoB are independent structures, we apply them sequentially: first, we convert the FFNs in the DiT into MoE layers, and then we group the MoE-equipped blocks into MoB.
\\[0.5em]
\textbf{Enhanced MoE Initialization.}
To replace the FFNs in the dense model with MoE layers, we initialize the experts in the MoE layer using the weights from the original FFNs. 
Inspired by pruning, we first compute and rank the importance of the weights to identify the most crucial ones in the model.  
Subsequently, we allocate the more important weights to the shared expert, through which all tokens pass, while distributing the less important weights among the normal experts.

Specifically, we use the first-order Taylor metric~\cite{molchanov2019importance} to compute the importance score \(\mathcal I_i\), which is formulated as
\begin{equation}
\mathcal I_i = \left| \frac{\partial \mathcal L}{\partial w_i} w_i \right|,
\end{equation}
where \( w_i \) represents the model weights, and \( \mathcal{L} \) is the loss function of diffusion model. 
We partition the MLP layers into \(r \times h\) segments based on the intermediate feature dimension. For each segment, we compute its accumulated importance score. We then select the top \(r_s \times h\) segments and reassemble them into a new MLP, which serves as the shared expert, while the remaining weights are evenly distributed among the normal experts.
When focusing solely on the shared expert, we treat it as a pruned model without gating or normal experts and apply knowledge distillation to enhance the shared expert, thereby improving the lower bound of the final sparse model.

For knowledge distillation (KD), we leverage both the output distill loss and block feature loss. 
The output distill loss is calculated as:
\begin{equation}
\mathcal{L}_{\text{distill}} = \mathbb{E}_{x_t,c,t}\| f_{\text{tea}}(x_t,c,t) - f_{\text{stu}}(x_t,c,t) \|_2^2,
\end{equation}
where \( f_{\text{tea}}(x_t,c,t) \) and \( f_{\text{stu}}(x_t,c,t) \) represent the outputs of the teacher and student models, respectively. 
The block feature loss is described as:
\begin{equation}
\mathcal{L}_{\text{feature}} = \sum_{l=1}^{L} w_l \| f_{\text{tea}}^{(l)}(x_t,c,t) - f_{\text{stu}}^{(l)}(x_t,c,t) \|_2^2,
\end{equation}
where \( L \) is the number of layers and \( w_l \) is the weight for each layer’s feature loss.
To prevent instability from varying feature scales and ensure effective learning in each layer, we incorporate methods from~\cite{zhang2024laptopdiff, minitron2024}. Specifically, we normalize the block feature loss using both the distillation loss and the L2-norm of the feature outputs from the teacher model. The feature weight is formulated as:
\begin{equation}
w_l = \frac{|\mathcal{L}_{\text{distill}}|} {|\mathcal{L}_{\text{feature}}^{(l)}|} \cdot \frac{\sum_{l=1}^L \| f_{\text{tea}}^{(l)}(x_t,c,t) \|_2}{L \cdot \| f_{\text{tea}, l}^{(l)}(x_t,c,t) \|_2}.
\end{equation}
\label{eq:reweight}
\\[0.5em]
\textbf{Dense-to-MoE Distillation.}
Subsequently, we activate the normal experts and the gating network, assembling them into the complete MoE layer. In this stage, we add several adjustments tailored to the MoE structure.
(1). Freezing shared experts: the shared experts had already been adequately trained in the previous stage, so to allow the distillation in this stage to focus on the normal experts and gating network and also improve training efficiency, we freeze the shared experts.
(2). Load balancing loss: To ensure that all normal experts are adequately trained and prevent the network collapse to a few experts, we utilize the load balancing loss~\cite{gshard}, which is formulated as
\[
\mathcal{L}_{\text{balance}} = \sum_{i=1}^{n} (\frac{n}{Tk} \sum_{t=1}^{T} I(t, i) \frac{1}{T} \sum_{t=1}^{T} g(x_i^{(t)}, t)),
\]
where \(T\) is the length of the token sequence, \(I(t, i)\) is the indicator function that denotes whether the \(t\)-th token selects expert \(i\).
Ultimately, the total loss for MoE distillation is given by
\begin{equation}
\mathcal{L}_{total} = \mathcal{L}_{\text{distill}} + \lambda_{\text{feature}} \mathcal{L}_{\text{feature}} + \lambda_{\text{balance}} \mathcal{L}_{\text{balance}},
\end{equation}
where \(\lambda_{\text{feature}} = 1\) and \(\lambda_{\text{balance}} = 10^{-2}\) are the weights for the block feature loss and the load balancing loss.
\\[0.5em]
\textbf{Group Feature Distillation.}
As shown in Fig.~\ref{fig:framework} (c), we design a group feature loss for MoB distillation. Although the blocks in DiT are grouped into MoBs, their relative order remains unchanged. 
We use the output features of the blocks in the original model that correspond to the last blocks of each MoB group as the teacher features, while the outputs of the MoB groups serve as the student features. We align them to improve the distillation process.
The group feature loss also follows the normalized feature weight as Eq.~\ref{eq:reweight}. 
Additionally, we freeze the isolated blocks that do not belong to any MoB group and apply load balancing loss. 

\section{Experiments}

\begin{table*}[h]
    \centering
    \renewcommand{\arraystretch}{1.2}
    \setlength{\tabcolsep}{4pt}
    \caption{Comprehensive comparison of our method with existing approaches in performance and efficiency, evaluated on H20 using grouped GEMM~\cite{gale2023megablocks}. All the samplings use FLUX.1 [dev] default settings with a guidance scale of 3.5, on 1024\(\times\)1024 resolution.}
    \small
    \scalebox{0.9}{
    \begin{tabular}{lc|ccc|cccccccc}
        \toprule
        \multirow{2}{*}{\textbf{Model}} & \multirow{2}{*}{\textbf{NFEs}} & \multirow{2}{*}{\makecell{\textbf{FLOPs} \\\textbf{(T)}$\downarrow$}} & \multirow{2}{*}{\makecell{\textbf{Activated} 
         \\ \textbf{Params (B)}}$\downarrow$} &  \multirow{2}{*}{\makecell{\textbf{Latency} \\\textbf{(s)}$\downarrow$}} & \multirow{2}{*}{\textbf{CLIP}$\uparrow$} & \multirow{2}{*}{\textbf{IR}$\uparrow$} & \multirow{2}{*}{\textbf{MPS}$\uparrow$} & \multirow{2}{*}{\textbf{GenEval}$\uparrow$} & \multirow{2}{*}{\textbf{DPG}$\uparrow$} & \multicolumn{3}{c}{\textbf{T2I-CompBench}} \\
\cmidrule(lr){11-13}
& & & & & & & & & & \textbf{B-VQA}$\uparrow$ & \textbf{UniDet}$\uparrow$ & \textbf{S-CoT}$\uparrow$ \\
        \midrule
        \textcolor{gray}{FLUX.1 [dev]} & \textcolor{gray}{28} & \textcolor{gray}{66.00} & \textcolor{gray}{11.90} & \textcolor{gray}{21.20} & \textcolor{gray}{32.24} & \textcolor{gray}{0.9656} & \textcolor{gray}{13.09} & \textcolor{gray}{0.6595} & \textcolor{gray}{83.42} & \textcolor{gray}{0.6401} & \textcolor{gray}{0.4262} & \textcolor{gray}{78.57} \\ 
        \textcolor{gray}{HyperFLUX~\cite{ren2024hypersd}} & \textcolor{gray}{8} & \textcolor{gray}{66.00} & \textcolor{gray}{11.90} & \textcolor{gray}{6.06} & \textcolor{gray}{32.19} & \textcolor{gray}{0.9859} & \textcolor{gray}{13.22} & \textcolor{gray}{0.6669} & \textcolor{gray}{83.25} & \textcolor{gray}{0.6270} & \textcolor{gray}{0.4375} & \textcolor{gray}{78.91} \\ 
        \hline
        FLUX.1-Lite~\cite{flux1-lite}    & 28 & 53.15 & 8.16 & 17.24 & \textbf{31.79} & \textbf{0.8380} & \textbf{12.87} & 0.5229 & 79.00 & 0.5471 & 0.3785 & 77.62 \\
        \textbf{FLUX.1-MoE-L} & 28 & 43.42 & 5.15 & 17.80 & 31.39 & 0.8011 & 12.60 & \underline{0.5702} & \underline{81.63} & \textbf{0.6032} & \underline{0.3899} & \textbf{78.28} \\
        \textbf{HyperFLUX-MoE-L} & 8 & 43.42 & 5.15 & 5.09 & \underline{31.50} & \underline{0.8257} & \underline{12.74} & \textbf{0.5917} & \textbf{81.76} & \underline{0.5959} & \textbf{0.3914} & \underline{78.13} \\ 
        \textbf{FLUX.1-MoE-M} & 28 & 35.70 & 4.01 & 14.37 & 30.77 & 0.5969 & 12.19 & 0.4758 & 76.26 & 0.4947 & 0.3400 & 77.50 \\ 
        \hline
        FLUX-Mini~\cite{flux_mini}   & 50 & \textbf{17.37} & 3.18 & 10.30 & 29.94 & 0.2151 & 11.18 & 0.3209 & 69.34 & 0.4209 & 0.2845 & 75.86 \\
        \textbf{FLUX.1-MoE-S} & 28 & 26.43 & 3.19 & 9.57 & 30.67 & 0.5942 & 12.06 & 0.4441 & 75.61 & 0.4958 & 0.3266 & 77.03 \\ 
        \textbf{FLUX.1-MoE-XS} & 28 & \underline{20.26} & \textbf{2.64} & 8.74 & 30.40 & 0.5076 & 11.80 & 0.4036 & 73.66 & 0.4734 & 0.3108 & 76.25 \\ 
        \textbf{HyperFLUX-MoE-XS} & 8 & \underline{20.26} & \textbf{2.64} & \textbf{2.50} & 30.70 & 0.5320 & 11.76 & 0.4117 & 74.53 & 0.4859 & 0.3036 & 76.88 \\
        \bottomrule
    \end{tabular}
    }
    \label{tab:model_comparison}    
\end{table*}

\subsection{Experimental Setup}
\textbf{Model Configuration.}
Based on FLUX.1 [dev] and HyperFLUX(8NFEs)~\cite{ren2024hypersd}, we conducted the main experiments by constructing four models with different sparsity levels (L, M, S, XS) with activation parameters ranging from 5.2 to 2.6B. (1). Level-L employs FFN-to-MoE, where the first three double-stream blocks remain unchanged, while the MLPs in the remaining 54 blocks are replaced with MoE layers (\( r_{share}=1, r_{normal}=0.25, n_{normal}=12, k=2 \)). Additionally, AdaLN compression is applied, reducing its channel dimension from 3072 to 256, lowering the total parameter count to 9B. (2). Level-M further introduces MoB sparsification, grouping 15 double-stream blocks into five MoB groups, each activating one block, thereby reducing the activated parameters of 10 double blocks. (3). Level-S and XS extend MoB sparsification to single-stream blocks, using three and five groups, respectively, with each MoB activating one of five blocks. This further reduces the activated parameters of 12 and 20 single blocks. 
\\[0.5em]
\textbf{Training Setting.} 
Our primary experiments are conducted using 32 NVIDIA A100 GPUs for multi-stage distillation, with a global batch size of 64. For training, we apply distillation using a combination of open-source datasets, including Laion-5B~\cite{schuhmann2022laion}, Coyo-700M~\cite{kakaobrain2022coyo-700m}, and JourneyDB~\cite{pan2023journeydb}.
\\[0.5em]
\textbf{Evaluation Metrics.} 
We evaluate text-to-image alignment, visual appeal, and human preference of the generated images using CLIP score~\cite{radford2021learning}, ImageReward score(IR)~\cite{xu2023imagereward}, and MPS reward score~\cite{zhang2024learning}. The evaluation is conducted on the MJHQ-30K dataset~\cite{li2024playground}, which consists of 30K images generated by Midjourney. Additionally, we employ GenEval~\cite{ghosh2023geneval}, DPG-Bench~\cite{hu2024ella}, and T2I-CompBench~\cite{huang2025t2icompbench++} to assess complex semantic alignment. For T2I-CompBench, we reorder the sub-metrics into B-VQA, UniDet, and S-CoT, as shown in Table~\ref{tab:model_comparison}.

\subsection{Main Results}
\textbf{Quantitative Comparison.}
In Tab.~\ref{tab:model_comparison}, we compare our models with existing pruning models distilled from FLUX.1 [dev]. FLUX.1-lite~\cite{flux1-lite} is an 8B model that removes 11 double stream blocks and FLUX-Mini~\cite{flux_mini} is a 3.2B model that reduces its depth to 5 double blocks and 10 single blocks.
Our principal observations are as follows:   
(1). Compared to FLUX.1-lite, our 5.2B acitvated parameter FLUX.1-MoE-L model achieves better performance in multiple benchmark, with \textbf{3B} fewer activated parameters and \textbf{20\%} fewer FLOPs.
(2). Under a 75\% compression rate of activated parameters, our method, whether it is the 3.19B activated FLUX.1-MoE-S or the 2.64B activated FLUX.1-MoE-XS, significantly outperforms the FLUX-Mini model, demonstrating the effectiveness of our approach at high compression rates.
(3). The results of HyperFLUX-MoE-L and XS show that our method maintains strong performance under step acceleration, achieving excellent results even with only 8 NFEs.

Furthermore, To comprehensively compare our proposed MoE approach with dense compression methods, we design two sets of comparative experiments.  
The first set evaluates MLP pruning~\cite{fang2023structural} and our MoE method, where \textbf{\(r\)} denotes the activated expand ratio of MLP or MoE after compression.  
The second set compares representative depth pruning methods~\cite{flux1-lite, kim2023bksdm} and MoB grouping under the same number of activated blocks.

\begin{table}[h]
    \centering
    \renewcommand{\arraystretch}{1.2}
    \setlength{\tabcolsep}{4pt}
    \small
    \caption{Comprehensive comparison of MLP Pruning and FFN-to-MoE, where \textbf{\(r_a\)} denotes the activated expand ratio of MLP or MoE after compression.}
    \label{tab:fair_compare}
    \scalebox{0.9}{
    \begin{tabular}{l|c|cccccc}
        \toprule
        \textbf{Methods} & \textbf{\(r_a\)} & \textbf{CLIP} & \textbf{IR} & \textbf{MPS} & \textbf{GenEval} & \textbf{DPG} \\
        \midrule
        DP~\cite{fang2023structural} & 1.5 & 30.89 & 0.4456 & 11.56 & 0.4113 & 72.23 \\
           & 2.0 & 31.31 & 0.6033 & 12.01 & 0.4888 & 77.53 \\
        \hline
        \textbf{MoE}  & 1.5 & \textbf{31.81} & \textbf{0.8368} & \textbf{12.74} & \textbf{0.5728} & \textbf{81.24} \\
        \bottomrule
    \end{tabular}
    \label{tab:moe_comparison}   
    }
\end{table}
\begin{table}[h]
    \centering
    \renewcommand{\arraystretch}{1.2}
    \setlength{\tabcolsep}{4pt}
    \small
    \caption{Comprehensive comparison of depth pruning and MoB grouping with equal numbers of activated transformer blocks, with \textbf{D} and \textbf{S} representing the number of activated double-stream and single-stream blocks, respectively.}
    \scalebox{0.9}{
    \begin{tabular}{l|cc|cccccccc}
        \toprule
        \textbf{Methods} & \textbf{D} & \textbf{S} & \textbf{CLIP} & \textbf{IR} & \textbf{MPS} & \textbf{GenEval} & \textbf{DPG} \\
        \midrule
        Lite~\cite{flux1-lite} & 9 & 38 & 31.48 & 0.7491 & 12.69 & 0.4919 & 77.72 \\
        BK~\cite{kim2023bksdm} & 9 & 38 & 31.20 & 0.6849 & 12.46 & 0.4789 & 76.39 \\
        \hline
        \textbf{MoB} & 9 & 38 & \textbf{31.59} & \textbf{0.7903} & \textbf{12.72} & \textbf{0.5396} & \textbf{78.86} \\
        \hline\hline
        Lite~\cite{flux1-lite}  & 9 & 26 & 26.64 & -1.1657 & 7.79 & 0.0926 & 41.62 \\
        BK~\cite{kim2023bksdm}   & 9 & 26 & 30.31 & 0.2587 & 11.13 & 0.3450 & 66.86 \\
        \hline
        \textbf{MoB} & 9 & 26 & \textbf{31.05} & \textbf{0.6541} & \textbf{12.27} & \textbf{0.4956} & \textbf{76.51} \\ 
        \bottomrule
    \end{tabular}
}
    \label{tab:mob_comparison}    
\end{table}

As shown in Tab.~\ref{tab:moe_comparison}, under the experimental setting where the activated expansion ratio is reduced from 4 to 1.5 (62.5\% compression), our MoE method outperforms Diff-Pruning~\cite{fang2023structural}.Further validation shows that MoE at 62.5\% compression still outperforms Diff-Pruning with a 50\% MLP compression.

Tab.~\ref{tab:mob_comparison} compares the performance of depth pruning and MoB grouping methods under the same activated parameters setting. We adopt a two-step distillation process: first reducing the model’s 19 double stream blocks to 9, then further compressing 38 single stream blocks to 26. Experimental results show that while the Lite method performs well when pruning only 10 double stream blocks, its performance drops significantly when an additional 12 single stream blocks are removed due to its single block distillation strategy. Although BK-SDM~\cite{kim2023bksdm}, which employs full-weight distillation, shows greater resilience than Lite, our MoB approach consistently achieves the best performance, highlighting its superior adaptability and effectiveness.

\subsection{Ablation Study}
For the sake of efficiency, we conduct the lightweight ablation studies at 512\(\times\)512 resolution, focusing only on the single-stream blocks in FLUX.1 [dev].
\\[0.5em]
\textbf{Trade-off Between Shared and Normal Experts.}
We examine two configurations to assess expert proportion impact: 1S12E2A (\(r_s=1, r_n=0.25, n=12, k=2\)) with a larger shared expert and fewer active normal experts, and 0.5S14E2A (\(r_s=0.5, r_n=0.25, n=14, k=4\)) with a smaller shared expert but more normal and activated experts.
Performance is evaluated in two stages: the initialization stage, using only shared experts, and the dense-to-MoE distillation stage, yielding the full sparse MoE model.  
Although both configurations maintain the same activated expand ratio, experimental results demonstrate that 1S12E2A achieves significantly better performance than 0.5S14E2A. Although reducing the size of the shared expert can increase model sparsity, it also increases the training difficulty at both stages, ultimately leading to a decline in overall performance. Finally, we selected 1S12E2A for the consideration of performance and computational efficiency.
\begin{table}[ht]
\centering
\small
\setlength{\tabcolsep}{4pt}
\scalebox{0.9}{
\begin{tabular}{cc|cccc}
\toprule
\textbf{Config} & \textbf{Infer Mode} & \textbf{CLIP} & \textbf{IR} & \textbf{MPS} & \textbf{GenEval} \\
\hline
\multirow{2}{*}{1S12E2A}  & Shared-only & 31.44   & 0.8028 & 12.18 & 0.6187 \\
                          & MoE & 31.52 & 0.8419 & 12.32 & 0.6331 \\
\hline
\multirow{2}{*}{0.5S14E4A}& Shared-only & 31.32 & 0.7591 & 11.84 & 0.6001 \\
                          & MoE & 31.48 & 0.8130 & 12.11 & 0.6270 \\ 

\bottomrule
\end{tabular}
}
\caption{Ablation study on the configurations of MoE. The notation \(x\)S\(y\)E\(z\)A represents a shared expert ratio of \(x\) with \(y\) normal experts (each at 0.25 ratio), where \(z\) experts are activated.} 
\label{tab:share_ablation}
\end{table}
\\[0.5em]
\textbf{Effects of Enhanced MoE Initialization and Distillation.}
We convert FFNs into MoEs through a two-stage process: enhanced initialization followed by MoE distillation. To evaluate the impact of key components, we conduct an ablation study, with results presented in Tab.~\ref{tab:moe_ablation}. The first two rows correspond to the first-stage distillation, where only shared experts are active, while the last two rows represent the second-stage results, where full MoE inference is performed.  
First, we compare the effectiveness of the Taylor metric against random weight splitting for shared expert distillation. The results demonstrate that the Taylor metric significantly improves the performance of shared experts.  
Next, we compare the full pipeline, where shared and normal experts are distilled separately, with a simplified strategy that distills them jointly. Experimental results indicate that independently optimizing shared and normal experts leads to superior performance.
\begin{table}[ht]
\centering
\small
\setlength{\tabcolsep}{4pt}
\scalebox{0.9}{
\begin{tabular}{cc|c|cccc}
\toprule
\multicolumn{2}{c|}{\textbf{MoE Init}} & \textbf{MoE} & \multirow{2}{*}{\textbf{CLIP}} & \multirow{2}{*}{\textbf{IR}} & \multirow{2}{*}{\textbf{MPS}} & \multirow{2}{*}{\textbf{GenEval}} \\
\cline{1-2}
\textbf{Taylor} & \textbf{Distill} & \textbf{Distill} \\
\hline
\checkmark & \checkmark &  & 31.44   & 0.8028 & 12.18 & 0.6187 \\
\ding{55} & \checkmark &  & (-0.02)   & (-0.159) & (-0.17) & (-0.0013) \\
\hline
\checkmark & \checkmark & \checkmark  & 31.52 & 0.8419 & 12.32 & 0.6331 \\
\checkmark & \ding{55} & \checkmark & (-0.04) & (-0.0204) & (-0.07) & (-0.0057) \\
\bottomrule
\end{tabular}
}
\caption{Ablation studies on enhanced MoE initialization and dense-to-MoE distillation.}
\label{tab:moe_ablation}
\end{table}
\\[0.5em]
\textbf{Effects of MoB Group Size and Group Feature Loss.}
To explore the optimal MoB grouping strategy, we evaluate three settings that each activate 4 out of 12 blocks:
(1) 4 groups of 3 blocks (1 active per group),
(2) 2 groups of 6 blocks (2 active per group), and
(3) a single group of 12 blocks (4 active).
The key difference lies in the routing mechanism, as each group operates independently with isolation between groups.  
Results indicate that a higher number of groups improves distillation performance, likely due to better feature alignment. More groups provide additional layers for supervision via group feature loss, enhancing overall alignment with the teacher model.
\begin{figure}[h]
\includegraphics[width=0.8\linewidth]{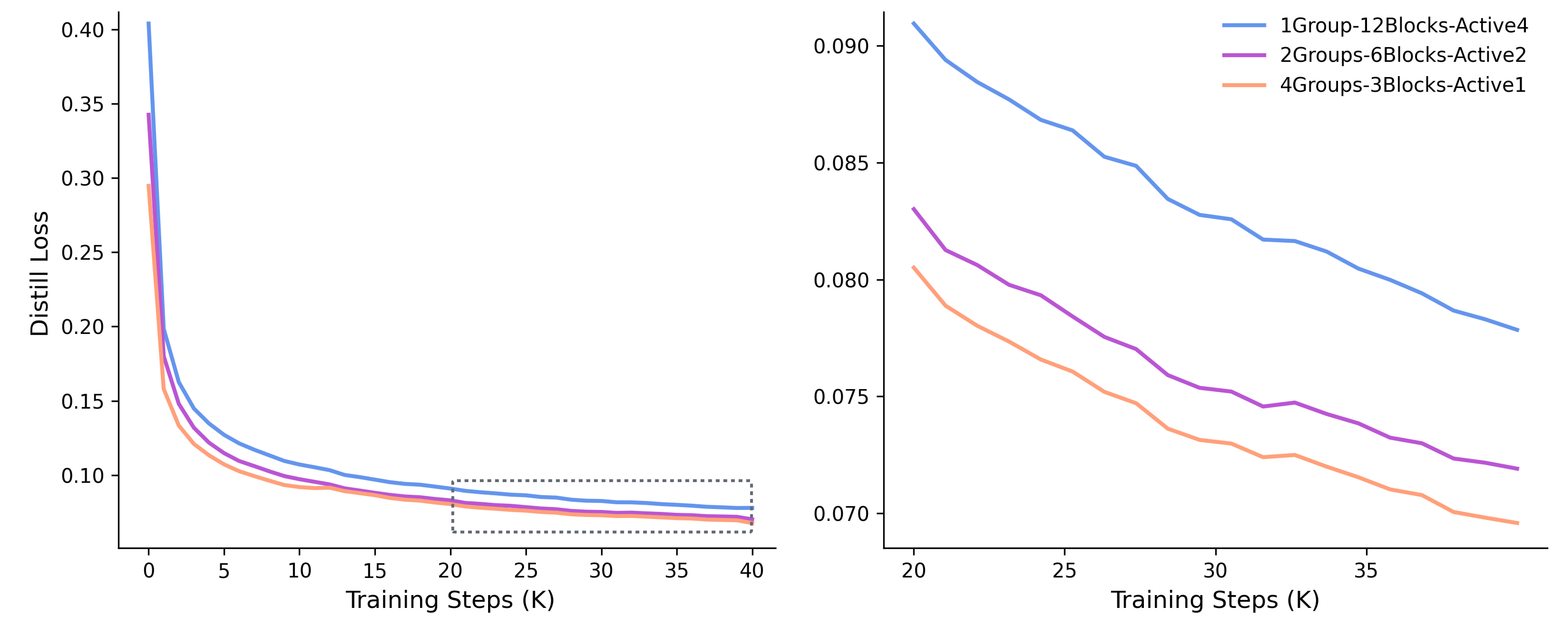}
\centering
\caption{Loss curves for different MoB group sizes during group feature distillation: 0–40K (left) and zoomed-in 20K–40K (right).}
\label{tab:block_mse_vis}
\end{figure}

\vspace{-1.0em} 
\subsection{Further Analysis}

\begin{figure}[h]
\includegraphics[width=0.9\linewidth]{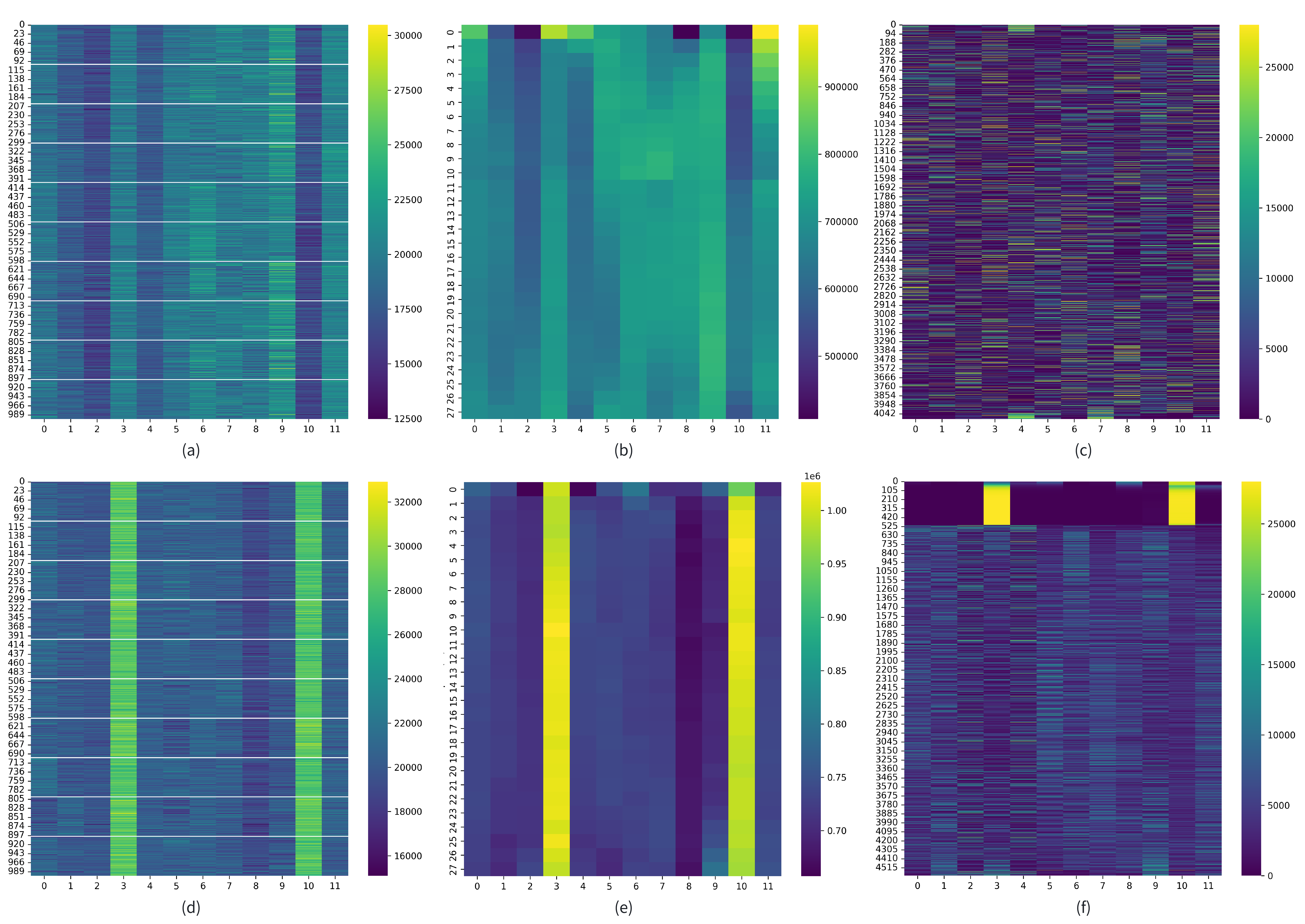}
\centering
\caption{Visualization of expert selection frequencies in the MoE layer. (a-c) show the expert selection frequencies of the MoE in the image branch of the 5th double stream block under prompt, timestep, and token conditions, respectively. (d-f) present the expert selection frequencies of the 5th single stream block under the same conditions. In (a) and (d), ten categories of prompts are separated by white horizontal lines.}
\label{img:expert}
\vspace{-1em}
\end{figure}

\textbf{Analysis of Expert Specialization.}
We conduct a comprehensive expert specialization analysis of our MoE model to explore its internal mechanisms. To compute the frequency of expert selection, we sample images using 1K prompts from the MJHQ-30K dataset, these 1K prompts are reorganized into 10 categories. 
In Fig.~\ref{img:expert}, we select two representative MoE layers. One is in the image branch
of the double stream block, one is in the single stream block.
We analyze expert specialization from three perspectives: prompts, timesteps, and tokens. 
We observe an interesting phenomenon in the single block, where some experts are selected more frequently than others, as shown in Fig.~\ref{img:expert} (d-e). This occurs because, in the single block, text and image tokens are concatenated. with the first 512 tokens corresponding to text and the remaining to image. As shown in Fig.~\ref{img:expert} (f), most text tokens are empty due to the length of prompts, resulting in convergent expert selection.
In contrast, the MoE layer in the image branch of the double block processes only image tokens, showing the following patterns: 
(1) From the prompt perspective, expert selection patterns are similar within the same prompt category, as seen in Fig.~\ref{img:expert} (a), where variations align with the white horizontal lines separating categories.  
(2) From the timestep perspective, expert selection changes throughout denoising, with more concentrated selection in the high-noise stage, especially at step 0 in Fig.~\ref{img:expert} (b).  
(3) From the token perspective, expert selection follows a spatial structure, showing a raster-like distribution in Fig.~\ref{img:expert} (c).

\begin{figure}[h]
\includegraphics[width=1.0\linewidth]{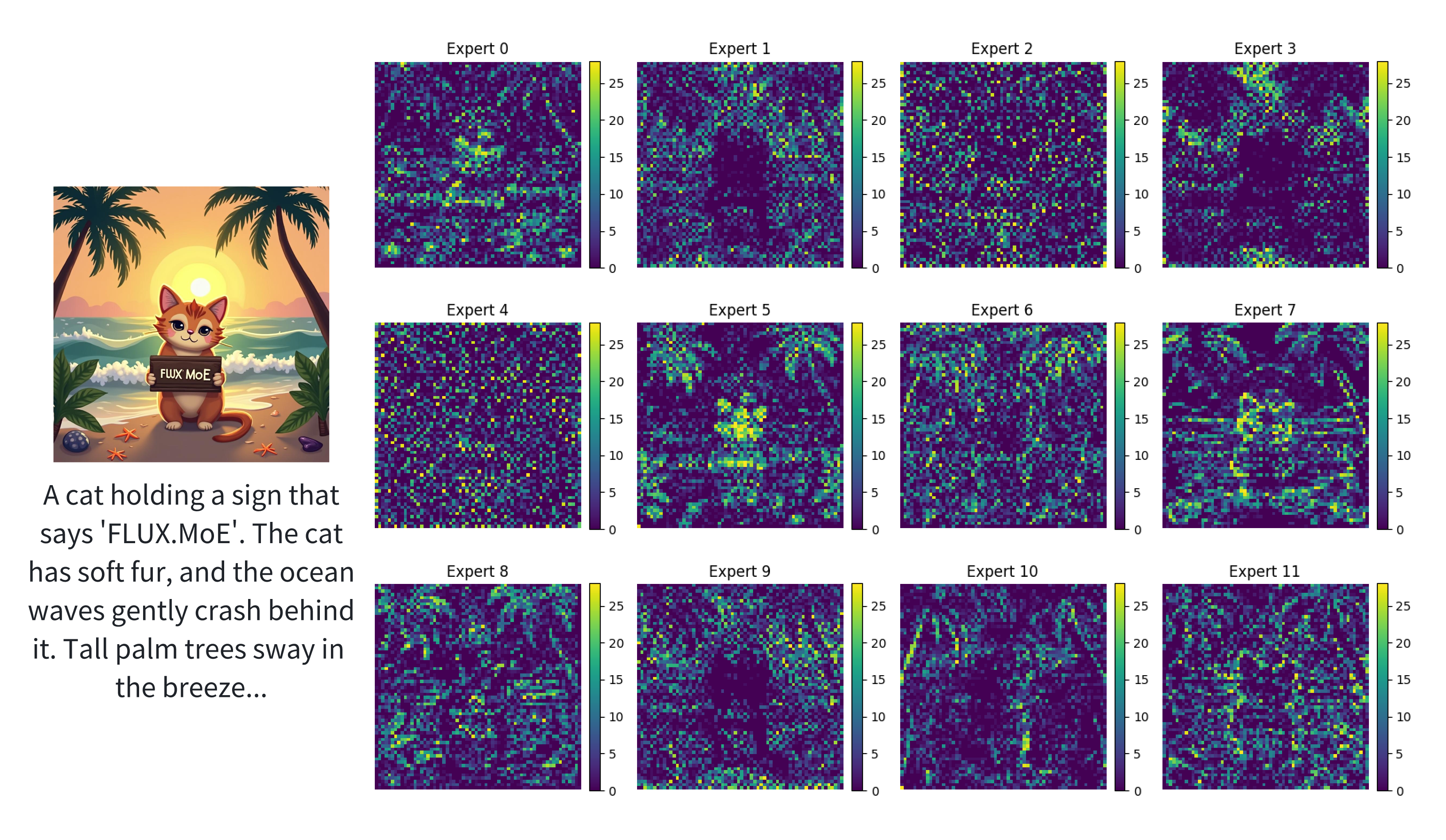}
\centering
\caption{Visualization of expert selection in the token dimension during single-image inference.}
\label{img:token_select}
\end{figure}
\begin{figure}[h]
\includegraphics[width=0.95\linewidth]{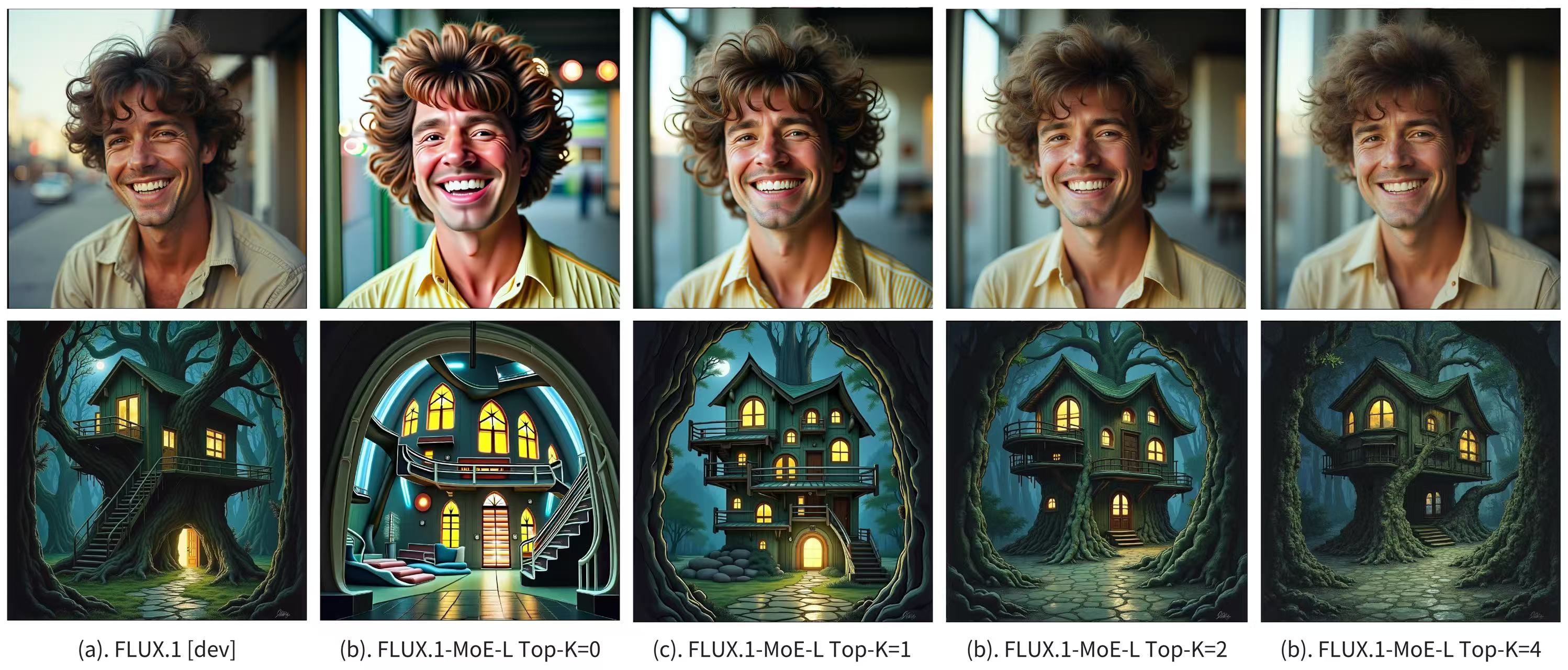}
\centering
\caption{Comparison with FLUX.1-MoE-L under different Top-K settings. Column (a) shows the original model, while (b-d) present FLUX.1-MoE-L results for Top-K = 0, 1, 2, and 4.}
\label{tab:mob_path}
\end{figure}

Additionally, we visualize expert selection at the token level using a single prompt in Fig.~\ref{img:token_select}. The results clearly indicate that different experts specialize in distinct spatial regions. These findings suggest that our sparse model distilled from a dense network offers strong network interpretability and aligns with previous observations~\cite{FeiDiTMoE2024}.
\\[0.5em]
\textbf{Dynamic TopK Activation for MoE.}
It is worth noting that due to our Dense-to-MoE distillation pipeline, the model retains basic generative capabilities even without activating normal experts, inherently supporting dynamic Top-K activation of normal experts.
To investigate this property, we evaluated inference performance under different Top-K settings without additional training, as presented in Fig.~\ref{tab:mob_path}. As observed, when the number of activated normal experts is reduced to zero, the generated outputs lose detail and exhibit higher color saturation. Conversely, increasing the number of activated experts improves detail and realism in the generated images.
The inherent sparsity and flexibility of MoEs open up new avenues for further exploration.

\vspace{-1.5em}

\section{Conclusion}  

In this paper, we propose Dense2MoE, a novel paradigm for efficient diffusion models that transforms a dense DiT into a sparse MoE. Our approach employs two sparsification strategies: replacing FFNs with MoE layers and grouping transformer blocks into MoB, along with a targeted distillation pipeline to restore performance.
Our method outperforms pruning-based techniques, compressing the 12B FLUX.1 [dev] model to 5.2B activation parameters while maintaining original performance. We believe our work will inspire further research on efficient generative models.

\section*{Acknowledgements}
This project is supported by the National Natural Science Foundation of China (12326618, U22A2095), the National Key Research and Development Program of China (2024YFA1011900), the Major Key Project of PCL (PCL2024A06), and the Project of Guangdong Provincial Key Laboratory of Information Security Technology (2023B1212060026).

{
    \small
    \bibliographystyle{ieeenat_fullname}
    \bibliography{main}

\begin{thebibliography}{44}
\providecommand{\natexlab}[1]{#1}
\providecommand{\url}[1]{\texttt{#1}}
\expandafter\ifx\csname urlstyle\endcsname\relax
  \providecommand{\doi}[1]{doi: #1}\else
  \providecommand{\doi}{doi: \begingroup \urlstyle{rm}\Url}\fi

\bibitem[Byeon et~al.(2022)Byeon, Park, Kim, Lee, Baek, and Kim]{kakaobrain2022coyo-700m}
Minwoo Byeon, Beomhee Park, Haecheon Kim, Sungjun Lee, Woonhyuk Baek, and Saehoon Kim.
\newblock Coyo-700m: Image-text pair dataset.
\newblock \url{https://github.com/kakaobrain/coyo-dataset}, 2022.

\bibitem[Chen et~al.(2023)Chen, Zhang, JAISWAL, Liu, and Wang]{chen2023sparse}
Tianlong Chen, Zhenyu Zhang, AJAY~KUMAR JAISWAL, Shiwei Liu, and Zhangyang Wang.
\newblock Sparse moe as the new dropout: Scaling dense and self-slimmable transformers.
\newblock In \emph{The Eleventh International Conference on Learning Representations}, 2023.

\bibitem[Daniel~Verdú(2024)]{flux1-lite}
Javier~Martín Daniel~Verdú.
\newblock Flux.1 lite: Distilling flux1.dev for efficient text-to-image generation.
\newblock 2024.

\bibitem[et~al.(2024)]{deepseekai2024deepseekv3technicalreport}
DeepSeek-AI et al.
\newblock Deepseek-v3 technical report, 2024.

\bibitem[Fang et~al.(2023)Fang, Ma, and Wang]{fang2023structural}
Gongfan Fang, Xinyin Ma, and Xinchao Wang.
\newblock Structural pruning for diffusion models.
\newblock In \emph{Advances in Neural Information Processing Systems}, 2023.

\bibitem[Fang et~al.(2024)Fang, Li, Ma, and Wang]{fang2024tinyfusion}
Gongfan Fang, Kunjun Li, Xinyin Ma, and Xinchao Wang.
\newblock Tinyfusion: Diffusion transformers learned shallow.
\newblock \emph{arXiv preprint arXiv:2412.01199}, 2024.

\bibitem[Fei et~al.(2024)Fei, Fan, Yu, Li, and Huang]{FeiDiTMoE2024}
Zhengcong Fei, Mingyuan Fan, Changqian Yu, Debang Li, and Jusnshi Huang.
\newblock Scaling diffusion transformers to 16 billion parameters.
\newblock \emph{arXiv preprint}, 2024.

\bibitem[Gale et~al.(2023)Gale, Narayanan, Young, and Zaharia]{gale2023megablocks}
Trevor Gale, Deepak Narayanan, Cliff Young, and Matei Zaharia.
\newblock Megablocks: Efficient sparse training with mixture-of-experts.
\newblock \emph{Proceedings of Machine Learning and Systems}, 5:\penalty0 288--304, 2023.

\bibitem[Ganjdanesh et~al.(2025)Ganjdanesh, Shirkavand, Gao, and Huang]{ganjdanesh2025aptp}
Alireza Ganjdanesh, Reza Shirkavand, Shangqian Gao, and Heng Huang.
\newblock Not all prompts are made equal: Prompt-based pruning of text-to-image diffusion models, 2025.

\bibitem[Ghosh et~al.(2023)Ghosh, Hajishirzi, and Schmidt]{ghosh2023geneval}
Dhruba Ghosh, Hannaneh Hajishirzi, and Ludwig Schmidt.
\newblock Geneval: An object-focused framework for evaluating text-to-image alignment.
\newblock \emph{Advances in Neural Information Processing Systems}, 36:\penalty0 52132--52152, 2023.

\bibitem[Gupta et~al.(2024)Gupta, Jaddipal, Prabhala, Paul, and Platen]{gupta2024progressive}
Yatharth Gupta, Vishnu~V. Jaddipal, Harish Prabhala, Sayak Paul, and Patrick~Von Platen.
\newblock Progressive knowledge distillation of stable diffusion xl using layer level loss, 2024.

\bibitem[Ho et~al.(2020)Ho, Jain, and Abbeel]{ho2020denoising}
Jonathan Ho, Ajay Jain, and Pieter Abbeel.
\newblock Denoising diffusion probabilistic models.
\newblock In \emph{Advances in Neural Information Processing Systems}, pages 6840--6851. Curran Associates, Inc., 2020.

\bibitem[Hu et~al.(2024)Hu, Wang, Fang, Fu, Cheng, and Yu]{hu2024ella}
Xiwei Hu, Rui Wang, Yixiao Fang, Bin Fu, Pei Cheng, and Gang Yu.
\newblock Ella: Equip diffusion models with llm for enhanced semantic alignment.
\newblock \emph{arXiv preprint arXiv:2403.05135}, 2024.

\bibitem[Huang et~al.(5555)Huang, Duan, Sun, Xie, Li, and Liu]{huang2025t2icompbench++}
Kaiyi Huang, Chengqi Duan, Kaiyue Sun, Enze Xie, Zhenguo Li, and Xihui Liu.
\newblock { T2I-CompBench++: An Enhanced and Comprehensive Benchmark for Compositional Text-to-Image Generation }.
\newblock \emph{IEEE Transactions on Pattern Analysis Machine Intelligence}, \penalty0 (01):\penalty0 1--17, 5555.

\bibitem[Jiang et~al.(2024)Jiang, Sablayrolles, Roux, Mensch, Savary, Bamford, Chaplot, de~las Casas, Hanna, Bressand, Lengyel, Bour, Lample, Lavaud, Saulnier, Lachaux, Stock, Subramanian, Yang, Antoniak, Scao, Gervet, Lavril, Wang, Lacroix, and Sayed]{jiang2024mixtralexperts}
Albert~Q. Jiang, Alexandre Sablayrolles, Antoine Roux, Arthur Mensch, Blanche Savary, Chris Bamford, Devendra~Singh Chaplot, Diego de~las Casas, Emma~Bou Hanna, Florian Bressand, Gianna Lengyel, Guillaume Bour, Guillaume Lample, Lélio~Renard Lavaud, Lucile Saulnier, Marie-Anne Lachaux, Pierre Stock, Sandeep Subramanian, Sophia Yang, Szymon Antoniak, Teven~Le Scao, Théophile Gervet, Thibaut Lavril, Thomas Wang, Timothée Lacroix, and William~El Sayed.
\newblock Mixtral of experts, 2024.

\bibitem[Kim et~al.(2023)Kim, Song, Castells, and Choi]{kim2023bksdm}
Bo-Kyeong Kim, Hyoung-Kyu Song, Thibault Castells, and Shinkook Choi.
\newblock Bk-sdm: A lightweight, fast, and cheap version of stable diffusion.
\newblock \emph{arXiv preprint arXiv:2305.15798}, 2023.

\bibitem[Labs(2024)]{flux2024}
Black~Forest Labs.
\newblock Flux.
\newblock \url{https://github.com/black-forest-labs/flux}, 2024.

\bibitem[Lee et~al.(2024)Lee, Park, Cho, Lee, and Hwang]{Lee2024koala}
Youngwan Lee, Kwanyong Park, Yoohrim Cho, Yong~Ju Lee, and Sung~Ju Hwang.
\newblock Koala: Empirical lessons toward memory-efficient and fast diffusion models for text-to-image synthesis.
\newblock In \emph{NeurIPS}, 2024.

\bibitem[Lepikhin et~al.(2020)Lepikhin, Lee, Xu, Chen, Firat, Huang, Krikun, Shazeer, and Chen]{gshard}
Dmitry Lepikhin, HyoukJoong Lee, Yuanzhong Xu, Dehao Chen, Orhan Firat, Yanping Huang, Maxim Krikun, Noam Shazeer, and Zhifeng Chen.
\newblock Gshard: Scaling giant models with conditional computation and automatic sharding.
\newblock \emph{CoRR}, abs/2006.16668, 2020.

\bibitem[Li et~al.(2024)Li, Kamko, Akhgari, Sabet, Xu, and Doshi]{li2024playground}
Daiqing Li, Aleks Kamko, Ehsan Akhgari, Ali Sabet, Linmiao Xu, and Suhail Doshi.
\newblock Playground v2.5: Three insights towards enhancing aesthetic quality in text-to-image generation.
\newblock 2024.

\bibitem[Lu et~al.(2022)Lu, Zhou, Bao, Chen, Li, and Zhu]{lu2022dpm}
Cheng Lu, Yuhao Zhou, Fan Bao, Jianfei Chen, Chongxuan Li, and Jun Zhu.
\newblock Dpm-solver: A fast ode solver for diffusion probabilistic model sampling in around 10 steps.
\newblock \emph{Advances in Neural Information Processing Systems}, 35:\penalty0 5775--5787, 2022.

\bibitem[Luo et~al.(2023)Luo, Tan, Huang, Li, and Zhao]{luo2023latent}
Simian Luo, Yiqin Tan, Longbo Huang, Jian Li, and Hang Zhao.
\newblock Latent consistency models: Synthesizing high-resolution images with few-step inference.
\newblock \emph{arXiv preprint arXiv:2310.04378}, 2023.

\bibitem[Molchanov et~al.(2019)Molchanov, Mallya, Tyree, Frosio, and Kautz]{molchanov2019importance}
Pavlo Molchanov, Arun Mallya, Stephen Tyree, Iuri Frosio, and Jan Kautz.
\newblock Importance estimation for neural network pruning.
\newblock In \emph{Proceedings of the IEEE/CVF conference on computer vision and pattern recognition}, pages 11264--11272, 2019.

\bibitem[Muralidharan et~al.(2024)Muralidharan, Sreenivas, Joshi, Chochowski, Patwary, Shoeybi, Catanzaro, Kautz, and Molchanov]{minitron2024}
Saurav Muralidharan, Sharath~Turuvekere Sreenivas, Raviraj Joshi, Marcin Chochowski, Mostofa Patwary, Mohammad Shoeybi, Bryan Catanzaro, Jan Kautz, and Pavlo Molchanov.
\newblock Compact language models via pruning and knowledge distillation.
\newblock In \emph{Advances in Neural Information Processing Systems (NeurIPS)}, 2024.

\bibitem[Pan et~al.(2023)Pan, Sun, Ge, Li, Duan, Wu, Zhang, Zhou, Qin, Wang, Dai, Qiao, and Li]{pan2023journeydb}
Junting Pan, Keqiang Sun, Yuying Ge, Hao Li, Haodong Duan, Xiaoshi Wu, Renrui Zhang, Aojun Zhou, Zipeng Qin, Yi Wang, Jifeng Dai, Yu Qiao, and Hongsheng Li.
\newblock Journeydb: A benchmark for generative image understanding, 2023.

\bibitem[Peebles and Xie(2023)]{peebles2023scalable}
William Peebles and Saining Xie.
\newblock Scalable diffusion models with transformers.
\newblock In \emph{Proceedings of the IEEE/CVF international conference on computer vision}, pages 4195--4205, 2023.

\bibitem[Radford et~al.(2021)Radford, Kim, Hallacy, Ramesh, Goh, Agarwal, Sastry, Askell, Mishkin, Clark, et~al.]{radford2021learning}
Alec Radford, Jong~Wook Kim, Chris Hallacy, Aditya Ramesh, Gabriel Goh, Sandhini Agarwal, Girish Sastry, Amanda Askell, Pamela Mishkin, Jack Clark, et~al.
\newblock Learning transferable visual models from natural language supervision.
\newblock In \emph{International conference on machine learning}, pages 8748--8763. PmLR, 2021.

\bibitem[Raposo et~al.(2024)Raposo, Ritter, Richards, Lillicrap, Humphreys, and Santoro]{raposo2024mixtureofdepths}
David Raposo, Sam Ritter, Blake Richards, Timothy Lillicrap, Peter~Conway Humphreys, and Adam Santoro.
\newblock Mixture-of-depths: Dynamically allocating compute in transformer-based language models, 2024.

\bibitem[Ren et~al.(2024)Ren, Xia, Lu, Zhang, Wu, Xie, Wang, and Xiao]{ren2024hypersd}
Yuxi Ren, Xin Xia, Yanzuo Lu, Jiacheng Zhang, Jie Wu, Pan Xie, Xing Wang, and Xuefeng Xiao.
\newblock Hyper-sd: Trajectory segmented consistency model for efficient image synthesis, 2024.

\bibitem[Rombach et~al.(2022)Rombach, Blattmann, Lorenz, Esser, and Ommer]{rombach2022high}
Robin Rombach, Andreas Blattmann, Dominik Lorenz, Patrick Esser, and Bj{\"o}rn Ommer.
\newblock High-resolution image synthesis with latent diffusion models.
\newblock In \emph{Proceedings of the IEEE/CVF conference on computer vision and pattern recognition}, pages 10684--10695, 2022.

\bibitem[Schuhmann et~al.(2022)Schuhmann, Beaumont, Vencu, Gordon, Wightman, Cherti, Coombes, Katta, Mullis, Wortsman, et~al.]{schuhmann2022laion}
Christoph Schuhmann, Romain Beaumont, Richard Vencu, Cade Gordon, Ross Wightman, Mehdi Cherti, Theo Coombes, Aarush Katta, Clayton Mullis, Mitchell Wortsman, et~al.
\newblock Laion-5b: An open large-scale dataset for training next generation image-text models.
\newblock \emph{Advances in neural information processing systems}, 35:\penalty0 25278--25294, 2022.

\bibitem[Shazeer et~al.(2017)Shazeer, Mirhoseini, Maziarz, Davis, Le, Hinton, and Dean]{shazeer2017outrageously}
Noam Shazeer, Azalia Mirhoseini, Krzysztof Maziarz, Andy Davis, Quoc Le, Geoffrey Hinton, and Jeff Dean.
\newblock Outrageously large neural networks: The sparsely-gated mixture-of-experts layer.
\newblock In \emph{International Conference on Learning Representations}, 2017.

\bibitem[Song et~al.(2020)Song, Meng, and Ermon]{song2020denoising}
Jiaming Song, Chenlin Meng, and Stefano Ermon.
\newblock Denoising diffusion implicit models.
\newblock \emph{arXiv preprint arXiv:2010.02502}, 2020.

\bibitem[Team(2024)]{qwen25}
Qwen Team.
\newblock Qwen2.5 technical report.
\newblock \emph{arXiv preprint arXiv:2412.15115}, 2024.

\bibitem[TencentARC(2023)]{flux_mini}
TencentARC.
\newblock Flux-mini.
\newblock \url{https://huggingface.co/TencentARC/flux-mini}, 2023.
\newblock Accessed: 2023-10-01.

\bibitem[Wu et~al.(2024)Wu, Zheng, He, and Yu]{wu2024parameter}
Haoyuan Wu, Haisheng Zheng, Zhuolun He, and Bei Yu.
\newblock Parameter-efficient sparsity crafting from dense to mixture-of-experts for instruction tuning on general tasks.
\newblock \emph{arXiv preprint arXiv:2401.02731}, 2024.

\bibitem[Xu et~al.(2023)Xu, Liu, Wu, Tong, Li, Ding, Tang, and Dong]{xu2023imagereward}
Jiazheng Xu, Xiao Liu, Yuchen Wu, Yuxuan Tong, Qinkai Li, Ming Ding, Jie Tang, and Yuxiao Dong.
\newblock Imagereward: Learning and evaluating human preferences for text-to-image generation.
\newblock \emph{Advances in Neural Information Processing Systems}, 36:\penalty0 15903--15935, 2023.

\bibitem[Xue et~al.(2023)Xue, Song, Guo, Liu, Zong, Liu, and Luo]{xue2023raphael}
Zeyue Xue, Guanglu Song, Qiushan Guo, Boxiao Liu, Zhuofan Zong, Yu Liu, and Ping Luo.
\newblock Raphael: Text-to-image generation via large mixture of diffusion paths.
\newblock \emph{Advances in Neural Information Processing Systems}, 36:\penalty0 41693--41706, 2023.

\bibitem[Yang et~al.(2024)Yang, Liu, Deng, Kim, Mei, Shen, and Chen]{yang2024158bitflux}
Chenglin Yang, Celong Liu, Xueqing Deng, Dongwon Kim, Xing Mei, Xiaohui Shen, and Liang-Chieh Chen.
\newblock 1.58-bit flux, 2024.

\bibitem[Yin et~al.(2024)Yin, Gharbi, Park, Zhang, Shechtman, Durand, and Freeman]{yin2024improved}
Tianwei Yin, Micha{\"e}l Gharbi, Taesung Park, Richard Zhang, Eli Shechtman, Fredo Durand, and William~T Freeman.
\newblock Improved distribution matching distillation for fast image synthesis.
\newblock In \emph{NeurIPS}, 2024.

\bibitem[Zhang et~al.(2024{\natexlab{a}})Zhang, Li, Chen, Xie, and Lu]{zhang2024laptopdiff}
Dingkun Zhang, Sijia Li, Chen Chen, Qingsong Xie, and Haonan Lu.
\newblock Laptop-diff: Layer pruning and normalized distillation for compressing diffusion models, 2024{\natexlab{a}}.

\bibitem[Zhang et~al.(2024{\natexlab{b}})Zhang, Wang, Wu, Li, Gao, Zhang, and Wang]{zhang2024learning}
Sixian Zhang, Bohan Wang, Junqiang Wu, Yan Li, Tingting Gao, Di Zhang, and Zhongyuan Wang.
\newblock Learning multi-dimensional human preference for text-to-image generation.
\newblock In \emph{Proceedings of the IEEE/CVF Conference on Computer Vision and Pattern Recognition}, pages 8018--8027, 2024{\natexlab{b}}.

\bibitem[Zhao et~al.(2024)Zhao, Han, Tang, Wang, Song, Huang, Wang, and You]{zhao2024dynamicdiffusiontransformer}
Wangbo Zhao, Yizeng Han, Jiasheng Tang, Kai Wang, Yibing Song, Gao Huang, Fan Wang, and Yang You.
\newblock Dynamic diffusion transformer, 2024.

\bibitem[Zhu et~al.(2024)Zhu, Guan, Liang, Chen, Liu, and Bai]{zhu2024moe}
Xingkui Zhu, Yiran Guan, Dingkang Liang, Yuchao Chen, Yuliang Liu, and Xiang Bai.
\newblock Moe jetpack: From dense checkpoints to adaptive mixture of experts for vision tasks.
\newblock \emph{Proceedings of Advances in Neural Information Processing Systems}, 2024.

\end{thebibliography}
}

\end{document}